\documentclass[10pt,twocolumn,letterpaper]{article}

\usepackage{iccv}
\usepackage{times}
\usepackage{epsfig}
\usepackage{graphicx}
\usepackage{amsmath}
\usepackage{amssymb}

\usepackage{enumitem}
\usepackage{booktabs}
\usepackage{float}
\usepackage{subfigure}
\usepackage{caption}
\usepackage{todonotes}
\usepackage{microtype}
\usepackage{array}
\usepackage{enumitem}

\newcolumntype{H}{>{\setbox0=\hbox\bgroup}c<{\egroup}@{}}

\usepackage{mathtools}

\newcommand{\fun}[3]{\ensuremath{#1\colon #2\mapsto #3}}

\DeclarePairedDelimiter\abs{\lvert}{\rvert}

\newcommand{\R}{\mathbb{R}}
\newcommand{\I}{\mathcal{I}}
\newcommand{\SO}{\ensuremath{\mathbf{SO}}}

\newcommand\blfootnote[1]{%
  \begingroup%
  \renewcommand\thefootnote{}\footnote{#1}%
  \addtocounter{footnote}{-1}%
  \endgroup%
}

\usepackage[pagebackref=true,breaklinks=true,letterpaper=true,colorlinks,bookmarks=false]{hyperref}

\iccvfinalcopy % *** Uncomment this line for the final submission

\def\iccvPaperID{5370} % *** Enter the ICCV Paper ID here

\ificcvfinal\fi
\begin{document}

%%%%%%%%% TITLE
\title{Equivariant Multi-View Networks}

\author{Carlos Esteves\textsuperscript{*},
  Yinshuang Xu\textsuperscript{*},
  Christine Allen-Blanchette,
  Kostas Daniilidis \\
  GRASP Laboratory, University of Pennsylvania \\
  {\tt\small \{machc,xuyin,allec,kostas\}@seas.upenn.edu}}

% \author{First Author\\
% Institution1\\
% Institution1 address\\
% {\tt\small firstauthor@i1.org}
% For a paper whose authors are all at the same institution,
% omit the following lines up until the closing ``}''.
% Additional authors and addresses can be added with ``\and'',
% just like the second author.
% To save space, use either the email address or home page, not both
% \and
% Second Author\\
% Institution2\\
% First line of institution2 address\\
% {\tt\small secondauthor@i2.org}
% }

\maketitle
\ificcvfinal\thispagestyle{empty}\fi

%%%%%%%%% ABSTRACT
\begin{abstract}
Several popular approaches to 3D vision tasks process multiple views of the input independently with deep neural networks pre-trained on natural images, achieving view permutation invariance through a single round of pooling over all views. We argue that this operation discards important information and leads to subpar global descriptors. In this paper, we propose a group convolutional approach to multiple view aggregation where convolutions are performed over a discrete subgroup of the rotation group, enabling, thus, joint reasoning over all views in an equivariant (instead of invariant) fashion, up to the very last layer. We further develop this idea to operate on smaller discrete homogeneous spaces of the rotation group, where a polar view representation is used to maintain equivariance with only a fraction of the number of input views. We set the new state of the art in several large scale 3D shape retrieval tasks, and show additional applications to panoramic scene classification.
\end{abstract}
\blfootnote{\textsuperscript{*} Equal contribution.}
\blfootnote{\phantom{\textsuperscript{*}} \url{http://github.com/daniilidis-group/emvn}}
\vspace{-0.5cm}
\begin{figure*}[h!]
  \centering
  \includegraphics[width=\textwidth]{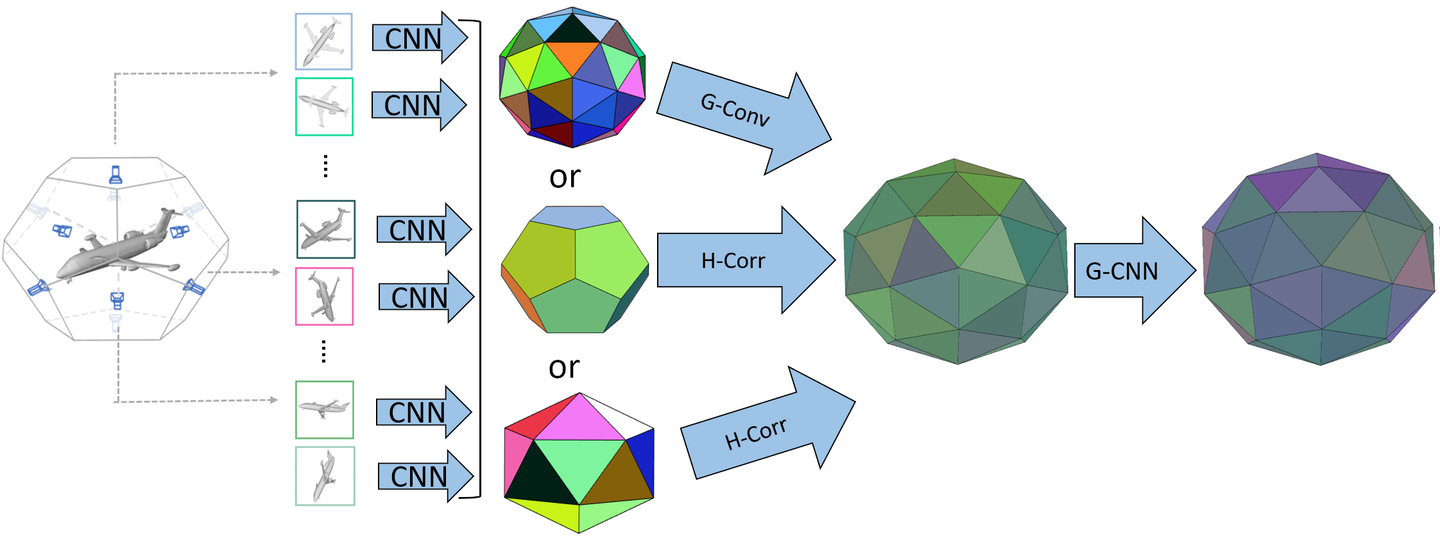}
  \caption{Our Equivariant Multi-View Network aggregates multiple views as functions on rotation groups that are processed via group convolutions.
    This guarantees equivariance to 3D rotations and allows jointly reasoning over all views, leading to superior shape descriptors.
    Vector-valued functions on the icosahedral group are shown on the pentakis dodecahedron, and functions on the corresponding homogeneous spaces (H-spaces) are shown on the dodecahedron and icosahedron.
    Each view is first processed by a CNN and resulting descriptors are associated with a group (or H-space) element.
    When views are identified with an H-space, the first operation is a correlation that lifts  features to the group.
    Once we have an initial representation on the group, a Group-CNN can be applied.}
    \label{structure}
  \end{figure*}

\section{Introduction}
The proliferation of large scale 3D datasets for objects \cite{wu20153d,shapenet} and whole scenes \cite{chang17_matter,dai17_scann} enables training of deep learning models producing global descriptors that can be applied to classification and retrieval tasks.

The first challenge that arises is how to represent the inputs.
Despite numerous attempts with volumetric \cite{wu20153d,voxnet}, point-cloud \cite{pointnet,simonovsky2017dynamic} and mesh-based \cite{Masci_2015_ICCV_Workshops,monti2017geometric} representations, using multiple views of the 3D input allows switching to the 2D domain where all the recent image based deep learning breakthroughs (\eg ~\cite{res}) can be directly applied, facilitating state of the art performance~\cite{multi,rotat}.

Multi-view (MV) based methods require some form of view-pooling, which can be
(1) pixel-wise pooling over some intermediate convolutional layer~\cite{multi},
(2) pooling over the final 1D view descriptors~\cite{su18_deeper_look_at_shape_class}, or
(3) combining the final logits~\cite{rotat}, which can be seen as independent voting.
These operations are usually invariant to view permutations.

Our key observation is that conventional view pooling is performed before any joint processing of the set of views and will inevitably discard useful features, leading to subpar descriptors.
We solve the problem by first realizing that each view can be associated with an element of the rotation group \SO(3), so the natural way to combine multiple views is as a function on the group.
A traditional CNN is applied to obtain view descriptors that compose this function.
We design a group-convolutional network (G-CNN, inspired by \cite{group}) to learn representations that are equivariant to transformations from the group.
This differs from the invariant representations obtained through usual view-pooling that discards  information.
We obtain invariant descriptors useful for classification and retrieval by pooling over the last G-CNN layer.
Our G-CNN has filters with localized support on the group and learns hierarchically more complex representations as we stack more layers and increase the receptive field.

We take advantage of the finite nature of multiple views and consider finite rotation groups like the icosahedral, in contrast with ~\cite{sph,learning} which operate on the continuous group.
To reduce the computational cost of processing one view per group element, we show that by considering views in canonical coordinates with respect to the group of in-plane dilated rotations (log-polar coordinates), we can greatly reduce the number of views and obtain an initial representation on a homogeneous space (H-space) that can be lifted via correlation, while maintaining equivariance.

We focus on 3D shapes but our model is applicable to any task where multiple views can represent the input, as demonstrated by an experiment on panoramic scenes.

Figure~\ref{structure} illustrates our model.
Our contributions are:
\begin{itemize}[topsep=2pt,itemsep=2pt]
\item We introduce a novel method of aggregating multiple views whether ``outside-in'' for 3D shapes or ``inside-out'' for panoramic views.
  Our model exploits the underlying group structure, resulting in equivariant features that are functions on the rotation group.
\item We introduce a way to reduce the number of views while maintaining equivariance, via a transformation to canonical coordinates of in-plane rotation followed by homogeneous space convolution.
\item We explore the finite rotation groups and homogeneous spaces and present a discrete G-CNN model on the largest group to date, the icosahedral group.
  We further explore the concept of filter localization for this group.
\item We achieve state of the art performance on multiple shape retrieval benchmarks, both in canonical poses and perturbed with rotations, and show applications to panoramic scene classification.
\end{itemize}

\vspace{0pt}\section{Related work}
% We divide the related work in (1) learned methods for 3D shape analysis and (2) equivariant representations.

%\subsection{3D object classification}
%For 3D volumeric data, there are several attempts proposed using CNNs with different representations of the data to address the problem of 3D object classification. Wu et al.\cite{shapenet}proposes 3D ShapeNets and Maturana \cite{voxnet}proposes VoxNet with fully-volumetric network. Qi et al.\cite{pointnet} train a convolutional network on the point-cloud dataset.Currently, the most successful approaches are view-based, operating in rendered views of the 3D object.Su et al\cite{multi} proposes to project the 3d mesh data to multi views which is processed on conventional 2D CNN. Qi et al.\cite{vam}proposes an alternate 3D CNN learning to project the volumetric representation to a 2D representation and then using a conventional 2D CNN architecture to learn the 2D processed representation.

\vspace{0pt}\paragraph{3D shape analysis}
Performance of 3D shape analysis is heavily dependent on the input representation.
The main representations are volumetric, point cloud and multi-view.

Early examples of volumetric approaches are~\cite{shapenet}, which introduced the ModelNet dataset and trained a 3D shape classifier using a deep belief network on voxel representations; and \cite{voxnet}, which presents a standard architecture with 3D convolutional layers followed by fully connected layers.

Su \etal~\cite{multi} realized that by rendering multiple views of the 3D input one can transfer the power of image-based CNNs to 3D tasks.
They show that a conventional CNN can outperform the volumetric methods even using only a single view of the input, while a multi-view (MV) model further improves the classification accuracy.

Qi \etal~\cite{vam} study volumetric and multi-view methods and propose improvements to both;
Kanezaki \etal~\cite{rotat} introduces an MV approach that achieves state-of-the-art classification performance by jointly predicting class and pose, though without explicit pose supervision.

GVCNN~\cite{gvcnn} attempts to learn how to combine different view descriptors to obtain a view-group-shape representation; they refer to arbitrary combinations of features as ``groups''.
This differs from our usage of the term  ``group'' which is the algebraic definition.

Point-cloud based methods~\cite{pointnet} achieve intermediate performance between volumetric and multi-view, but are much more efficient computationally.
While meshes are arguably the most natural representation and widely used in computer graphics, only limited success has been achieved with learning models operating directly on them \cite{Masci_2015_ICCV_Workshops,monti2017geometric}.

% The ModelNet dataset is small compared to current image datasets, and classification performance on it is considered saturated.
In order to better compare 3D shape descriptors we will focus on the retrieval performance.
Recent approaches show significant improvements on retrieval: You \etal~\cite{you2018pvnet} combines point cloud and MV representations;
Yavartanoo \etal~\cite{yavartanoo18_spnet} introduces multi-view stereographic projection; and
Han \etal \cite{han2019seqviews2seqlabels} implements a recurrent MV approach.

We also consider more challenging tasks on rotated ModelNet and SHREC'17~\cite{savva2017shrec} retrieval challenge which contains rotated shapes.
The presence of arbitrary rotations motivates the use of equivariant representations.

\vspace{0pt}\paragraph{Equivariant representations}
A number of workarounds have been introduced to deal with 3D shapes in arbitrary orientations.
Typical examples are training time rotation augmentation and/or test time voting \cite{vam} and
learning an initial rotation to a canonical pose \cite{pointnet}.
The view-pooling in~\cite{multi} is invariant to permutations of the set of input views.

A principled way to deal with rotations is to use representations that are equivariant by design.
There are mainly three ways to embed equivariance into CNNs.
The first way is to constrain the filter structure, which is similar to Lie generator based approach ~\cite{1992,1998}.
Worral \etal~\cite{harmonic} take advantage of circle harmonics to have both translational and 2D rotational equivariance into CNNs.
Similarly, Thomas \etal~\cite{tensor} introduce a tensor field to keep translational and rotational equivariance for 3D point clouds.

The second way is through a change of coordinates;
\cite{polar,henriques2017warped} take the log-polar transform of the input and transfer rotational and scaling equivariance about a single point to translational equivariance.

The third way is to make use of an equivariant filter orbit.
Cohen and Welling propose group convolution (G-CNNs) with the square rotation group~\cite{group}, later extended to the hexagon~\cite{hex}.
Worrall and Brostow~\cite{cubenet} proposed CubeNet using Klein’s Four-group on 3D voxelized data. Winkels \etal~\cite{cohencube} implement 3D group convolution on Octahedral symmetry group  for volumetric CT images.
Cohen \etal~\cite{cohen19_gauge_equiv_convol_networ_icosah_cnn} very recently considered functions on the icosahedron, however their convolutions are on the cyclic group and not on the icosahedral as ours.
Esteves \etal~\cite{learning} and Cohen \etal~\cite{sph} focus on the infinite group \SO(3), and use the spherical harmonic transform for the exact implementation of the spherical convolution or correlation.
The main issue with these approaches is that the input spherical representation does not capture the complexity of an object's shape;
they are also less efficient and face bandwidth challenges.
% Even if the spherical input would represent a 2D view rendering for every point on the sphere, that would be a too dense and overcomplete representation, and would require an unnecessary computation in producing a dense sphere of 2D renderings for every testing input. A multiple view representation of a 3D object is powerful because few views can capture the whole object and be quite discriminative as well.

% Group convolutional networks were introduced by Cohen and Welling \cite{group}, where they consider small groups of 4 rotations and 0/1 reflections. Worrall and Brostow \cite{cubenet} extend this idea to the symmetries of the cube.

%   On the other hand, since uneven discretization is a problem of \SO(3), we can use its 'even' finite subgroup. We use the icosahedral group, which is the biggest 'even' finite group in \SO(3). 2D multi-view rendering would be a good method to obtain the group descriptor. Our method takes every single view into convolution rather than do pooling, and we use an initialization method to show that \textbf{MVCNN\cite{multi} is just a special situation in our method.} What's more, we use the rotation-equivariant representation, therefore we don't need to rotate the views to augment data like MVCNN\cite{multi}.

\section{Preliminaries}
We seek to leverage symmetries in data.
A {\em symmetry} is an operation that preserves some structure of an object.
If the object is a discrete set with no additional structure, each operation can be seen as a permutation of its elements.

The term \emph{group} is used in its classic algebraic definition of a set with an operation satisfying the closure, associativity, identity, and inversion properties.
A transformation group like a permutation is the ``missing link between abstract group and the notion of symmetry'' ~\cite{miller72}.

We refer to \emph{view} as an image taken from an oriented camera.
This differs from \emph{viewpoint} that refers to the optical axis direction, either \emph{outside-in} for a moving camera pointing at a fixed object, or \emph{inside-out} for a fixed camera pointing at different directions.
Multiple \emph{views} can be taken from the same \emph{viewpoint}; they are related by in-plane rotations.

\vspace{-2pt}\paragraph{Equivariance}
Representations that are equivariant by design are an effective way to exploit symmetries.
Consider a set $\mathcal{X}$ and a transformation group $G$. For any $g\in G$, we can define group action applied on the set,
% \begin{equation}\label{ET1}
$\mathcal{T}^{\mathcal{X}}_{g}:\mathcal{X}\rightarrow \mathcal{X}$,
% \end{equation}
which has property of homomorphism,
% \begin{equation}\label{ET2}
$\mathcal{T}^{\mathcal{X}}_{g}\mathcal{T}^{\mathcal{X}}_{h}=\mathcal{T}^{\mathcal{X}}_{gh}$.
% \end{equation}
Consider a map $\Phi:\mathcal{X}\rightarrow\mathcal{Y}$.
We say $\Phi$ is equivariant to $G$ if
\begin{equation}
\Phi(\mathcal{T}^{\mathcal{X}}_{g}(x))=\mathcal{T}^{\mathcal{Y}}_{g}(\Phi(x)),\quad \forall  x\in \mathcal{X}, g\in G.
\end{equation}
In the context of CNNs, $\mathcal{X}$ and $\mathcal{Y}$ are sets of input and feature representations, respectively.
This definition encompasses the case when $\mathcal{T}^{\mathcal{Y}}_{g}$ is the identity, making $\Phi$ invariant to $G$ and discarding information about $g$.
In this paper, we are interested in non-degenerate cases that preserve information.

\vspace{-2pt}\paragraph{Convolution on groups}
We represent multiple views as a functions on a group and seek equivariance to the group, so group convolution (G-Conv) is the natural operation for our method.
Let us recall planar convolution between \fun{f,h}{\R^2}{\R}, which is the main operation of CNNs:
\begin{equation}
  (f * h)(y) = \int_{x\in\R^2} f(x)h(y-x)\,dx.
\end{equation}
It can be seen as an operation over the group of translations on the plane, where the group action is addition of coordinate values;
it is easily shown to be equivariant to translation.
This can be generalized to any group $G$ and \fun{f,h}{G}{\R},
\begin{equation}
  (f * h)(y) = \int_{g\in G} f(g)h(g^{-1}y)\,dg, \label{eq:gconv}
\end{equation}
which is equivariant to group actions from $G$.

% In this work, we are mainly interested on convolutions on discrete rotation groups, where (\ref{eq:gconv}) can be written as follows,
% \begin{equation}
%   (F * H)(y) = \sum_{g\in G} F(g)H(g^{-1}y).
% \end{equation}
% \todo[inline]{discrete group conv.}

\vspace{-2pt}\paragraph{Convolution on homogeneous spaces}
\label{sec:homogeneous}
For efficiency, we may relax the requirement of one view per group element and consider only one view per element of a homogeneous space of lower cardinality.
For example, we can represent the input on the 12 vertices of the icosahedron (an H-space), instead of on the 60 rotations of the icosahedral group.

A homogeneous space $\mathcal{X}$ of a group $G$ is defined as a space where $G$ acts transitively: for any $x_1,x_2 \in \mathcal{X}$, there exists $g\in G$ such that $x_2=gx_1$.

% For example, there exists $g \in \SO(3)$ that relates any two points on the sphere $\mathbf{S}^2$, so the sphere is a homogeneous space of \SO(3).

Two convolution-like operations can be defined between functions on homogeneous spaces \fun{f,h}{\mathcal{X}}{\R}:
\begin{align}
  (f * h)(y) &= \int_{g\in G} f(g\eta)h(g^{-1}y)\,dg, \label{eq:hconv} \\
  (f \star h)(g) &= \int_{x\in \mathcal{X}} f(gx)h(x)\,dx, \label{eq:hcorr}
\end{align}
where $\eta \in \mathcal{X}$ is an arbitrary canonical element.
We denote (\ref{eq:hconv}) ``homogeneous space convolution'' (H-Conv), and (\ref{eq:hcorr}) ``homogeneous space correlation'' (H-Corr).
Note that convolution produces a function on the homogeneous space $\mathcal{X}$ while correlation lifts the output to the group $G$.

We refer to~\cite{kondor18_gener_equiv_convol_neural_networ,cohen18_gener_theor_equiv_cnns_homog_spaces} for expositions on group and homogeneous space convolution in the context of neural networks.

% \todo[inline]{give examples: faces/vertices of ico and their dual; faces of truncated ico}
\vspace{-2pt}\paragraph{Finite rotation groups}
\label{sec:rotgroups}
Since our representation is a finite set of views that can be identified with rotations, we will deal with finite subgroups of the rotation group \SO(3).
A finite subgroup of \SO(3) can be the cyclic group $\mathcal{C}_k$ of multiples of $2\pi/k$, the dihedral group $\mathcal{D}_k$ of symmetries of a regular $k$-gon, the tetrahedral, octahedral, or  icosahedral group~\cite{artin}.

% To compare with the state of the art performance of planar/azimuthal rotations, we will use
% the 12-element cyclic group, which contains 12 rotations in the plane, whose rotation degrees are $\{30i|i \in \mathbb{Z}_{12} \}$. It is an abelian group isomorphic to $\mathbb{Z}_{12}$ that will be denoted with $\boldsymbol{C}_{12}$.

Our main results are on the icosahedral group $\I$, the 60-element non-abelian group of symmetries of the icosahedron (illustrated in the supplementary material).
The symmetries can be divided in sets of rotations around a few axes.
For example, there are 5 rotations around each axis passing through vertices of the icosahedron or 3 rotations around each axis passing through its faces centers.
% We denote an element in the group as $g_i$.
% The icosahedral rotation group is isomorphic to the group of even permutations of a set, called  the alternating group $\mathcal{A}_{5}$ which in general contains $n!/2$ elements consistent with our 60 rotations for $n=5$.

% The group includes one identity rotation, 4 rotations along each of 6 axes linking 2 opposite vertices, 2 rotations along each of 10 axes linking 2 opposite central points of faces and 1 rotation along each of 15 axes linking 2 opposite middle points on edges.
% To illustrate the binary operations in these groups, we show their Cayley tables in the supplementary material.

\vspace{-2pt}\paragraph{Equivariance via canonical coordinates}
\label{sec:canoncoords}
Some configurations produce views that are related by in-plane rotations.
We leverage this to reduce the number of required views by obtaining rotation invariant view descriptors through a change to canonical coordinates followed by a CNN.

Segman \etal \cite{1992} show that changing to a canonical coordinate system allows certain transformations of the input to appear as translations of the output.
For the group of dilated rotations on the plane (isomorphic to $\SO(2) \times \R^+$), canonical coordinates are given by the log-polar transform.
% \todo[inline]{log polar transform}

Since planar convolutions are equivariant to translation, converting an image to log-polar and applying a CNN results in features equivariant to dilated rotation, which can be pooled to invariant descriptors on the last layer \cite{polar,henriques2017warped}.

\section{Method}
Our first step is to obtain $\abs{G}$ views of the input where each view $x_i$ is associated with a group element $g_i \in G$%
\footnote{Alternatively, we can use $\abs{\mathcal{X}}$ views for a homogeneous space $\mathcal{X}$ as shown in \ref{sec:fewerviews}.}.
Each view is fed to a CNN $\Phi_1$, and the 1D descriptors extracted from the last layer (before projection into the number of classes) are combined to form a function on the group \fun{y}{G}{\R^n}, where $y(g_i) = \Phi_1(x_i)$.
A group convolutional network (G-CNN) $\Phi_2$ operating on $G$ is then used to process $y$, and global average pooling on the last layer yields an invariant descriptor that is used for classification or retrieval.
Training is end-to-end.
Figure~\ref{structure} shows the model.

The MVCNN with late-pooling from \cite{rotat}, which outperforms the original \cite{multi}, is a special case of our method where $\Phi_2$ is the identity and the descriptor is $y$ averaged over $G$.

% reordering stuff
\begin{figure*}[ht!]
 \centering
 \includegraphics[width=\linewidth]{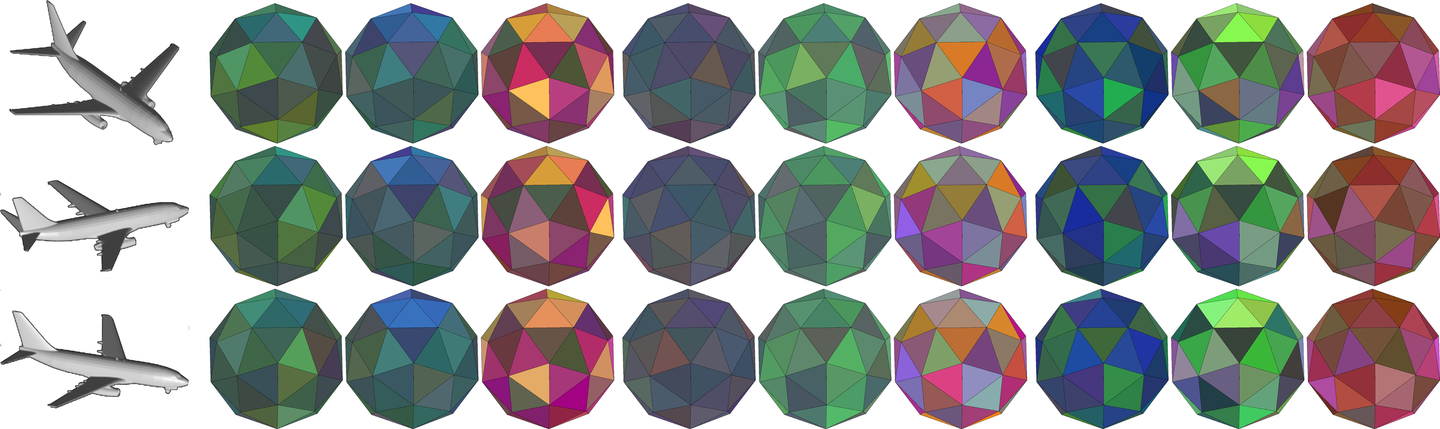}
 \caption{Features learned by our method are visualized on the pentakis dodecahedron, which has icosahedral symmetry so its 60 faces can be identified with elements of the discrete rotation group $\I$.
   Columns show learned features from different channels/layers.
   The first two rows are related by a rotation of 72 deg in $\I$.
   Equivariance is exact in this case, as can be verified by the feature maps rotating around the polar axis (notice how the top 5 cells shift one position).
   The first and third row are related by a rotation of 36 deg around the same axis, which is in the midpoint between two group elements.
 Equivariance is approximate in this case, and features are a mixture of the two above.}
\label{fig:fmaps}
\end{figure*}

\subsection{View configurations}
\label{sec:viewconf}
There are several possible view configurations of icosahedral symmetry, basically consisting of vertices or faces of solids with the same symmetry.
Two examples are associating viewpoints with faces/vertices of the icosahedron, which are equivalent to the vertices/faces of its dual, the dodecahedron.
% To generate views based on icosahedron vertices, we place virtual cameras at the 12 vertices of the icosahedron and rotate by $2k\pi/5,\, k\in\{0,1,2,3,4\}$, around the axis passing through each camera.
These configurations are based on platonic solids, which guarantee a uniform distribution of viewpoints.
By selecting viewpoints from the icosahedron faces, we obtain 20 sets of 3 views that differ only by 120 deg in plane rotations; we refer to this configuration as $20\times 3$.
Similarly, using the dodecahedron faces we obtain the $12\times 5$ configuration.

In the context of 3D shape analysis, multiple viewpoints are useful to handle self-occlusions and ambiguities.
Views that are related by in-plane rotations are redundant in this sense, but necessary to keep the group structure.

\begin{figure}[htb]
 \centering
 \includegraphics[width=\linewidth]{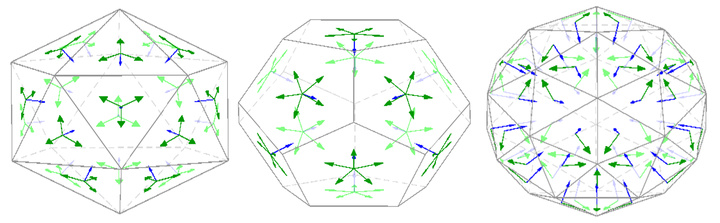}
\caption{Outside-in camera configurations considered.
  Left to right: $20\times 3$, $12\times 5$, and $60\times 1$.
  Blue arrows indicate the optical axis and green, the camera up direction.
  Object is placed at the intersection of all optical axes.
  Only the $60\times 1$ configuration avoids views related by in-plane rotations.}
\label{fig:camera}
\end{figure}

To minimize redundancy, we propose to associate viewpoints with the 60 vertices of the truncated icosahedron (which has icosahedral symmetry).
There is a single view per viewpoint in this configuration.
This is not a uniformly spaced distribution of viewpoints, but the variety is beneficial.
Figure~\ref{fig:camera} shows some view configurations we considered.
% We show how the equivariance manifests as view permutation in the supplementary material.

Note that our configurations differ from both the 80-views from \cite{multi} and 20 from \cite{rotat} which are not isomorphic to any rotation group.
Their 12-views configuration is isomorphic to the more limited cyclic group.

\subsection{Group convolutional networks}

The core of the group convolutional part of our method is the discrete version of (\ref{eq:gconv}).
A group convolutional layer with $c_i,c_j$ channels in the input and output and nonlinearity $\sigma$ is then given by
\begin{align}
  f_j^{\ell+1}(y) = \sigma\left( \sum_{i=1}^{c_i}\sum_{g\in G} f_i^\ell(g)h_{ij}(g^{-1}y) \right),
  % F^{\ell+1}(i,n) = \sum\sum F^{\ell}(j,m) H(C_{j'i},m,n)
  \label{eq:gconvd}
\end{align}
where $f_i^\ell$ is the channel $i$ at layer $\ell$ and $h_{ij}$ is the filter between channels $i$ and $j$, where $1 \leq j \leq c_j$.
This layer is equivariant to actions of $G$.

Our most important results are on the icosahedral group $\I$ which has 60 elements and is the largest discrete subgroup of the rotation group \SO(3).
To the best of our knowledge, this is the largest group ever considered in the context of discrete G-CNNs.
Since $\I$ only coarsely samples \SO(3), equivariance to arbitrary rotations is only approximate.
Our results show, however, that the combination of invariance to local deformations provided by CNNs  and exact equivariance by G-CNNs is powerful enough to achieve state of the art performance in many tasks.

When considering the group $\I$, inputs to $\Phi_2$ are $60 \times n$ where $n$ is the number of channels in the last layer of $\Phi_1$ (n=512 for ResNet-18).
There are $c_i \times c_j$ filters per layer each with up to the same cardinality of the group.

We can visualize both filters and feature maps as functions on the faces of the pentakis dodecahedron, which is the dual polyhedron of the truncated icosahedron.
It has icosahedral symmetry and 60 faces that can be identified with elements of the group.
The color of the face associated with $g_i\in\I$ reflects $f(g_i)$, which is vector valued.
Figure~\ref{fig:fmaps} shows some equivariant feature maps learned by our method.

\subsection{Equivariance with fewer views}
\label{sec:fewerviews}
As illustrated in Figure~\ref{fig:camera}, the icosahedral symmetries can be divided in sets of rotations around a few axes.
If we arrange the cameras such that they lie on these axes, images produced by each camera are related by in-plane rotations.

As shown in Section~\ref{sec:canoncoords}, converting one image to canonical coordinates can transform in-plane rotations in translations.
We'll refer to converted images as ``polar images''.
Since fully convolutional networks can produce translation-invariant descriptors, by applying them to polar images we effectively achieve invariance to in-plane  rotations~\cite{polar,henriques2017warped}, which makes only one view per viewpoint necessary.
These networks require circular padding in the angular dimension.

When associating only a single view per viewpoint, the input is on a space of points instead of a group of rotations%
\footnote{They are isomorphic for the $60\times 1$ configuration.}.
In fact, the input is a function on a homogeneous space of the group; concretely, for the view configurations we consider, it is on the icosahedron or dodecahedron vertices.

We can apply discrete versions of convolution and correlation on homogeneous spaces as defined in Section~\ref{sec:homogeneous}:
\begin{align}
  ^*f_j^{\ell+1}(y) &= \sigma\left( \sum_{i=1}^{c_i}\sum_{g\in G} f_i^\ell(g\eta)h_{ij}(g^{-1}y) \right),\\
  {^\star}f_j^{\ell+1}(g) &= \sigma\left( \sum_{i=1}^{c_i}\sum_{x\in \mathcal{X}} f_i^\ell(gx)h_{ij}(x) \right).
\end{align}

\begin{figure}[htb]
  \centering
  \includegraphics[width=\linewidth]{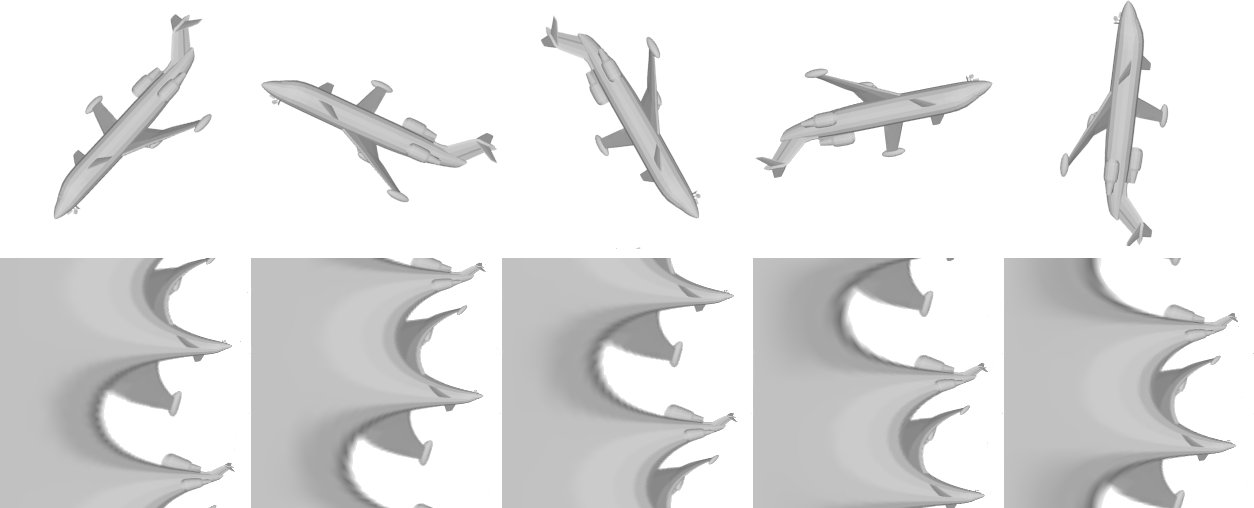}
  \caption{One subset of in-plane related views from the $12\times 5$ configuration and correspondent polar images.
    Note how the polar images are related by circular vertical shifts so their CNN descriptors are approximately invariant to the in-plane rotation.
    There are 12 such subsets for the $12\times 5$ configuration and 20 for the $20\times 3$; this allows us to maintain equivariance with 12 or 20 views instead of 60.}
 \label{fig:polar}
\end{figure}

The benefit of this approach is that since it uses 5x (3x) fewer views when starting from the $12\times 5$ ($20\times 3$) configuration, it is roughly  5x (3x) faster as most of the compute is done before the G-CNN.
The disadvantage is that learning from polar images can be challenging.
Figure~\ref{fig:polar} shows one example of polar images produced from views.

When inputs are known to be aligned (in canonical pose), an equivariant intermediate representation is not necessary; in this setting, we can use the same method to reduce the number of required views, but without the polar transform.

\subsection{Filter localization}
G-CNN filters are functions on $G$, which can have up to $\abs{G}$ entries.
Results obtained with deep CNNs during the past few years show the benefit from limited support filters (many architectures use $3\times 3$ kernels throughout).
The advantages are two-fold: (1) convolution with limited support is computationally more efficient, and (2) it allows learning of hierarchically more complex features as layers are stacked.
Inspired by this idea, we introduce localized filters for discrete G-CNNs%
\footnote{Localization for the continuous case was introduced in \cite{learning}.}.
For a filter \fun{h}{G}{\R}, we simply choose a subset $S$ of $G$ that is allowed to have nonzero filter values while $h(G - S)$ is set to zero.
Since $S$ is a fixed hyperparameter, we can compute~(\ref{eq:gconvd}) more efficiently:
\begin{align}
  f_j^{\ell+1}(y) = \sigma\left( \sum_{i=1}^{c_i}\sum_{g\in S} f_i^\ell(yg^{-1})h_{ij}(g) \right).
\end{align}

To ensure filter locality, it is desirable that elements of $S$ are close to each other in the manifold of rotations.
The 12 smallest rotations in $\I$ are of $72$ deg.
We therefore choose $S$ to contain the identity and a number of $72$ deg rotations.

\begin{figure}[htb]
  \centering
  \includegraphics[width=\linewidth]{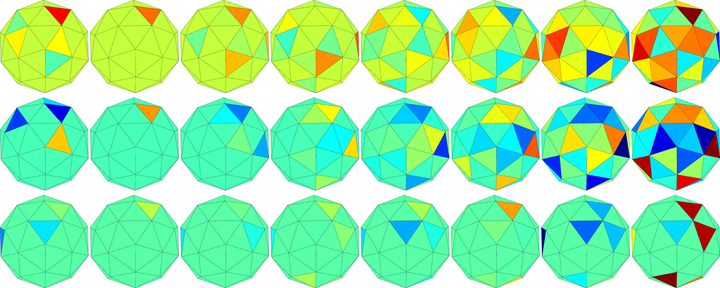}
  \caption{Localized filters and their receptive fields as we stack more layers.
    First column shows the filter, second the input, and others are results of stacking group convolutions with the same filter.
    Top row filter has 12 nonzero elements; middle and bottom have 5.
    The support for the bottom row contains elements of a 12 element subgroup, so its receptive field cannot cover the full input space.}
\label{fig:localized}
\end{figure}

One caveat of this approach is that we need to make sure $S$ spans $G$, otherwise the receptive field will not cover the whole input no matter how many layers are stacked, which can happen if $S$ belongs to a subgroup of $G$ (see Figure~\ref{fig:localized}).
In practice this is not a challenging condition to satisfy; for our heuristic of choosing only $72$ deg rotations we only need to guarantee that at least two are around different axes.

\section{Experiments}
We evaluate on 3D shape classification, retrieval and scene classification, and include more comparisons and an ablation study in the supplementary material.
First, we discuss the architectures, training procedures, and datasets.
\vspace{-2pt}\paragraph{Architectures}
We use a ResNet-18~\cite{res} as the view processing network $\Phi_1$, with weights initialized from ImageNet~\cite{imagenet_cvpr09} pre-training.
The G-CNN part contains 3 layers with 256 channels and 9 elements on its support (note that the number of parameters is the same as one conventional $3\times 3$ layer).
We project from 512 to 256 channels so the number of parameters stay close to the baseline.
When the method in Section~\ref{sec:fewerviews} is used to reduce the number of views, the first G-Conv layer is replaced by a H-Corr.

Variations of our method are denoted Ours-X, and Ours-R-X.
The R suffix indicate retrieval specific features, that consist of (1) a triplet loss%
\footnote{Refer to the supplementary material for details.} and (2) reordering the retrieval list so that objects classified as the query's predicted class come first.
Before reordering, the list is sorted by cosine distance between descriptors.
For SHREC'17, choosing the number N of retrieved objects is part of the task -- in this case we simply return all objects classified as the query's class.

For fair assessment of our contributions, we implement a variation of MVCNN, denoted MVCNN-M-X for X input views, where the best-performing X is shown.
MVCNN-M-X has the same view-processing network, training procedure and dataset as ours;
the only difference is that it performs pooling over view descriptors instead of using a G-CNN.

\vspace{-2pt}\paragraph{Training} We train using SGD with Nesterov momentum as the optimizer.
For ModelNet experiments we train for 15 epochs, and 10 for SHREC'17.
Following~\cite{he18_bag_trick_image_class_with}, the learning rate linearly increases from 0 to $lr$ in the first epoch, then decays to zero following a cosine quarter-cycle.
When training with 60 views, we set the batch size to 6, and $lr$ to $0.0015$.
This requires around 11~Gb of RAM.
When training with 12 or 20 views, we linearly increase both the batch size and $lr$.

Training our 20-view model on ModelNet40 for one epoch takes $\approx 353s$ on an NVIDIA 1080
Ti, while the corresponding MVCNN-M takes $\approx 308s$.
Training RotationNet \cite{rotat} for one epoch under same conditions takes $\approx 1063s$.

\vspace{-2pt}\paragraph{Datasets}
We render $12\times 5$, $20\times 3$ and $60\times 1$ camera configurations (Section~\ref{sec:viewconf}) for ModelNet and the ShapeNet SHREC'17 subset, for both rotated and aligned versions.
For the aligned datasets, where equivariance to rotations is not necessary, we fix the camera up-vectors to be in the plane defined by the object center, camera and north pole.
This reduces the number of views from $12\times 5$ to 12 and from $20\times 3$ to 20.
For the rotated datasets, all renderings have 60 views and follow the group structure.
Note that the rotated datasets are not limited to the discrete group and contain continuous rotations from $\SO(3)$.
We observe that the $60\times 1$ configuration performs best so those are the numbers shown for ``Ours-60''.
For the experiment with fewer views, we chose 12 from $12\times 5$ and 20 from $20\times 3$ that are converted to log-polar coordinates (Section~\ref{sec:fewerviews}).
For the scene classification experiment, we sample 12 overlapping views from panoramas.
No data augmentation is performed.

\subsection{SHREC'17 retrieval challenge}
The SHREC'17 large scale 3D shape retrieval challenge~\cite{savva2017shrec} utilizes the ShapeNet Core55~\cite{shapenet} dataset and has two modes: ``normal'' and ``perturbed'' which correspond to ``aligned'' and ``rotated'' as we defined in Section~\ref{sec:modelnet}.
The challenge was carried out in 2017 but there has been recent interest on it, especially on the ``rotated'' mode~\cite{sph,learning,kondor2018clebsch}.
% The dataset is divided in train/val/test sets with 35,764/5,133/10,265 entries, respectively.
% We combine training and validation sets to produce the final results.

Table~\ref{tab:shrec} shows the results.
N is the number of retrieved elements, which we choose to be the objects classified as the same class as the query.
The Normalized Discounted Cumulative Gain (NDGC) score uses ShapeNet subclasses to measure relevance between retrieved models.
Methods are ranked by the mean of micro (instance-based) and macro (class-based) mAP.
Several extra retrieval metrics are included in the supplementary material.
Only the best performing methods are shown; we refer to~\cite{savva2017shrec} for more results.

Our model outperforms the state of the art for both modes even without the triplet loss, which, when included, increase the margins.
We consider this our most important result, since it is the largest available 3D shape retrieval benchmark and there are numerous published results on it.

\begin{table}[ht!]
  \centering
  \small
    \begin{tabular}{@{}lrrHHHrrrHHHrr@{}}
      \toprule
                                               &           & \phantom{} & \multicolumn{5}{c}{micro} & \phantom{} & \multicolumn{5}{c}{macro}                                                                        \\
      \cmidrule{2-2} \cmidrule{4-8} \cmidrule{10-14}
                  Method                       & score     &            & P@N                       & R@N        & F1@N      & mAP       & G@N       &  & P@N       & R@N       & F1@N      & mAP       & G@N       \\
      \midrule
                  RotatNet~\cite{rotat}        & 67.8      &            & 81.0                      & 80.1       & 79.8      & 77.2      & 86.5      &  & 60.2      & 63.9      & 59.0      & 58.3      & 65.6      \\
                  ReVGG~\cite{savva2017shrec}  & 61.8      &            & 76.5                      & 80.3       & 77.2      & 74.0      & 82.8      &  & 51.8      & 60.1      & 51.9      & 49.6      & 55.9      \\
                  DLAN~\cite{furuya2016deep} & 57.0      &            & 81.8                      & 68.9       & 71.2      & 66.3      & 76.2      &  & 61.8      & 53.3      & 50.5      & 47.7      & 56.3      \\
                  % MVCNN-12~\cite{multi}      & 65.1      &            & 77.0                      & 77.0       & 76.4      & 73.5      & 81.5      &  & 57.1      & 62.5      & 57.5      & 56.6      & 64.0      \\
                 MVCNN-M-12                    & 69.1      &            & 83.1                      & 77.9       & 79.4      & 74.9      & 83.8      &  & \it{66.8} & 68.4      & 65.2      & 63.2      & 70.3      \\
                  Ours-12                      & 70.7      &            & 83.1                      & 80.5       & 81.1      & 77.7      & 86.3      &  & 65.3      & 68.7      & 64.8      & 63.6      & 70.8      \\
                  Ours-20                      & 71.4      &            & \it{83.6}                 & \it{80.8}  & \it{81.5} & \it{77.9} & \it{86.8} &  & 66.4      & \it{70.1} & 65.9      & 64.9      & 71.9      \\
                  Ours-60                      & \it{71.7} &            & \bf{84.0}                 & 80.5       & 81.4      & 77.8      & 86.4      &  & \bf{67.1} & \bf{70.7} & \bf{66.6} & \bf{65.6} & \bf{72.3} \\
                  Ours-R-20                    & \bf{72.2} &            & \it{83.6}                 & \bf{81.7}  & \bf{82.0} & \bf{79.1} & \bf{87.5} &  & \it{66.8} & 69.9      & \it{66.1} & \it{65.4} & \bf{72.3} \\
      \midrule
                  DLAN~\cite{furuya2016deep}   & 56.6      &            & \bf{81.4}                 & 68.3       & 70.6      & 65.6      & 75.4      &  & \bf{60.7} & 53.9      & 50.3      & 47.6      & 56.0      \\
                  ReVGG~\cite{savva2017shrec}  & 55.7      &            & 70.5                      & \bf{76.9}  & 71.9      & \it{69.6} & 78.3      &  & 42.4      & 56.3      & 43.4      & 41.8      & 47.9      \\
                  % MVCNN-80~\cite{multi}      & 45.1      &            & 63.2                      & 61.3       & 61.2      & 53.5      & 65.3      &  & 40.5      & 48.4      & 41.6      & 36.7      & 45.9      \\
                 MVCNN-M-60                    & 57.5      &            & 77.7                      & 67.6       & 71.1      & 64.1      & 75.9      &  & 55.7      & 56.9      & 53.5      & 50.9      & 59.7      \\
                  Ours-12                      & 58.1      &            & 76.1                      & 70.0       & 72.0      & 66.4      & 76.7      &  & 54.6      & 55.7      & 52.6      & 49.8      & 58.6      \\
                  Ours-20                      & 59.3      &            & 76.4                      & 70.5       & 72.4      & 66.9      & 77.0      &  & 54.6      & 58.0      & 53.7      & 51.7      & 60.2      \\
                  Ours-60                      & \it{62.1} &            & \it{78.7}                 & 72.9       & \it{74.7} & \it{69.6} & \it{79.6} &  & 57.6      & \it{60.1} & \it{56.3} & \it{54.6} & \it{63.0} \\
                  Ours-R-60                    & \bf{63.5} &            & \it{78.7}                 & \it{75.0}  & \bf{75.9} & \bf{71.8} & \bf{81.1} &  & \it{58.3} & \bf{60.6} & \bf{56.9} & \bf{55.1} & \bf{63.3} \\
      \bottomrule
    \end{tabular}
  \caption{SHREC'17 retrieval results.
    Top block: aligned dataset; bottom: rotated.
    We show mean average precision (mAP) and normalized discounted cumulative gain (G).
    % Methods are ranked by the micro and macro mAP average (``score'').
    We set the new state of the art by a large margin.
    Even our 12-view model outperforms the baseline,
    which shows the potential of keeping equivariance with fewer views.}
  \label{tab:shrec}
\end{table}

\subsection{ModelNet classification and retrieval}
\label{sec:modelnet}
We evaluate 3D shape classification and retrieval on variations of ModelNet~\cite{wu20153d}.
In order to compare with most publicly available results, we evaluate on ``aligned'' ModelNet, and use all available models with the original train/test split (9843 for training, 2468 for test).
% Note that some methods~\cite{wu20153d,multi} use only a subset of ModelNet40 with 3183 models for training and 800 for test.
% We include results for this subset on the supplementary material. \todo{!}
We also evaluate on the more challenging ``rotated'' ModelNet40, where each instance is perturbed with a random rotation from \SO(3).

Tables~\ref{tab:modelnet} and \ref{tab:rotmodelnet} show the results.
We show only the best performing methods and refer to the ModelNet website\footnote{\url{http://modelnet.cs.princeton.edu}} for complete leaderboard.
Classification performance is given by accuracy (acc) and retrieval by the mean average precision (mAP).
Averages are over instances.
We include class-based averages on the supplementary material.

We outperform the retrieval state of the art for both ModelNet10 and ModelNet40, even without retrieval-specific features.
When including such features (triplet loss and reordering by class label), the margin increases  significantly.

We focus on retrieval and do not claim state of the art on classification, which is held by RotationNet~\cite{rotat}.
While ModelNet retrieval was not attempted by~\cite{rotat}, the SHREC'17 retrieval was, and we show significantly better performance on it (Table~\ref{tab:shrec}).

\begin{table}
  \centering
    \begin{tabular}{@{}llllll@{}}
      \toprule
                                             & \multicolumn{2}{c}{M40 (aligned)} & \phantom{abc} & \multicolumn{2}{c}{M10 (aligned)} \\
      \cmidrule{2-3} \cmidrule{5-6}
                                             & acc                               & mAP           &  & acc        & mAP               \\
      \midrule
      MVCNN-12~\cite{multi}                  & 90.1                              & 79.5          &  & -          & -                 \\
      SPNet~\cite{yavartanoo18_spnet}        & 92.63                             & 85.21         &  & \it{97.25} & 94.20             \\
      PVNet~\cite{you2018pvnet}              & 93.2                              & 89.5          &  & -          & -                 \\
      SV2SL~\cite{han2019seqviews2seqlabels} & 93.40                             & 89.09         &  & 94.82      & 91.43             \\
      PANO-ENN \cite{sfikas2018ensemble}     & \bf{95.56}                        & 86.34         &  &	96.85      & 93.2              \\
      % RotationNet~\cite{rotat}             & \bf{97.37}                        & -             &  & \bf{98.46} & -                 \\
      MVCNN-M-12                             & 94.47                             & 89.13         &  & 96.33      & 93.54             \\
      Ours-12                                & 94.51                             & \it{91.82}    &  & 96.33      & 95.30             \\
      Ours-20                                & \it{94.69}                        & 91.42         &  & \bf{97.46} & \it{95.74}        \\
      Ours-60                                & 94.36                             & 91.04         &  & 96.80      & 95.25             \\
      Ours-R-12                              & 94.67                             & \bf{93.56}    &  & 96.78      & \bf{96.18}        \\
      \bottomrule%
    \end{tabular}
    \caption{Aligned ModelNet classification and retrieval.
      We only compare with published retrieval results.
      We achieve state of the art retrieval performance even without retrieval-specific model features.
      This shows that our view aggregation is useful even when global equivariance is not necessary.}
  \label{tab:modelnet}
\end{table}

% | 03-08 03:50 | mvgcnn60-m40ti3d-r18-3l-12s-nfc1-2        | 85.20 (85.27)   |    90.96 |   89.09 |    82.64 |   78.11 | 0.3403 |
%                 mvgcnn60-m40ti3d-r18-3l-12s-nfc1-triplet-1| 88.19 (88.19)   |    91.17 |   88.96 |    88.34 |   84.31 | 0.3356 |
% | 03-08 04:47 | mvcnn60-m40ti3d-r18-nopre-vlavg-3         | 82.56 (82.56)   |    90.68 |   88.13 |    78.18 |   73.25 | 0.3118 |

% | 03-10 21:18 | mvgcnn60-m40d3d-h12-polarc-r18-3l-12s-h1st-2  | 82.90 (83.22)   |    89.27 |   86.44 |    80.38 |   75.52 | 0.4051 |
% | 03-10 18:43 | mvgcnn60-m40i3d-h20-polarc-r18-3l-12s-h1st-2  | 83.12 (83.12)   |    89.50 |   87.66 |    80.25 |   75.06 | 0.3989 |

% mvgcnn60-m40d3d-h12-r18-3l-9s-256-h1st-1 & 88.50 & 79.58\\
% mvgcnn60-m40i3d-h20-r18-3l-9s-256-h1st-1 & 89.98 & 80.73\\
% mvgcnn60-m40ti3d-r18-3l-9s-256-1 & 91.00 & 82.61\\
% mvgcnn60-m40ti3d-r18-3l-9s-256-triplet-1 & 91.08 & 88.57\\

\begin{table}
  \centering
    \begin{tabular}{@{}lll@{}}
      \toprule
                               & \multicolumn{2}{c}{M40 (rotated)} \\
      \cmidrule{2-3}
                               & acc        & mAP                  \\
      \midrule
      MVCNN-80~\cite{multi}    & 86.0       & -                    \\
      RotationNet~\cite{rotat} & 80.0       & 74.20                \\
      Spherical CNN~\cite{sph} & 86.9       & -                    \\
      MVCNN-M-60               & 90.68      & 78.18                \\
      Ours-12                  & 88.50      & 79.58                \\
      Ours-20                  & 89.98      & 80.73                \\
      Ours-60                  & \it{91.00} & \it{82.61}           \\
      Ours-R-60                & \bf{91.08} & \bf{88.57}           \\
      \bottomrule
    \end{tabular}
    \caption{Rotated ModelNet40 classification and retrieval.
      Note that gap between ``Ours'' and ``MVCNN-M'' is much larger than in the aligned dataset,
      which demonstrates the advantage of our equivariant representation.}
  \label{tab:rotmodelnet}
\end{table}

\subsection{Scene classification}
We have shown experiments for object-centric configurations (outside-in),
but our method is also applicable to camera-centric configurations (inside-out),
which is demonstrated on the Matterport3D~\cite{chang17_matter} scene classification from panoramas task.
We sample multiple overlapping azimuthal views from the panorama and apply our model over the cyclic group of 12 rotations, with a filter support of 6.
Table~\ref{tab:sceneclass} shows the results; the full table with accuracies per class and input samples are in the supplementary material.

The MV approach is superior to operating directly on panoramas because (1) it allows higher overall resolution while sharing weights across views, and (2) views match the scale of natural images so pre-training is better exploited.
Our MVCNN-M outperforms both baselines, and our proposed model outperforms it, which shows that the group structure is also useful in this setting.
In this task, our representation is equivariant to azimuthal rotations; a CNN operating directly on the panorama has the same property.%, while the MVCNN-M is invariant.

\begin{table}
  \centering
    \begin{tabular}{lrrrr}
      \toprule
      & single \cite{chang17_matter} & pano \cite{chang17_matter} & MV-M-12 & Ours-12 \\
      \midrule
      acc [\%] & 33.3 & 41.0 & 51.9 & 53.8\\
      \bottomrule
    \end{tabular}
    \caption{Scene classification class-based accuracy.}
  \label{tab:sceneclass}
\end{table}%
% \subsection{Ablation?}
% \subsection{Pose estimation}
\subsection{Discussion}
Our model shows state of the art performance on multiple 3D shape retrieval benchmarks.
We argue that the retrieval problem is more appropriate to evaluate shape descriptors because it requires a complete rank of similarity between models instead of only a class label.
%; besides, classification results for ModelNet40 seem to be saturated.

Our results for aligned datasets show that the full set of 60 views is not necessary and may be even detrimental in this case; but even when equivariance is not required, the principled view aggregation with G-Convs is beneficial, as direct comparison between MVCNN-M and our method show.
For rotated datasets, results clearly show that performance increases with the number of views, and that the aggregation with G-Convs brings huge improvements.

Interestingly, our MVCNN-M baseline outperforms many competing approaches.
The differences with respect to the original MVCNN~\cite{multi} are (1) late view-pooling, (2) use of ResNet, (3) improved rendering, and (4) improved learning rate schedule.
These significant performance gains were also observed in~\cite{su18_deeper_look_at_shape_class}, and attest to the representative potential of multi-view representations.

One limitation is that our feature maps are equivariant to discrete rotations only, and while classification and retrieval performance under continuous rotations is excellent, for tasks such as continuous pose estimation it may not be.
Another limitation is that we assume views to follow the group structure, which may be difficult to achieve for real images.
Note that this is not a problem for 3D shape analysis, where we can render any arbitrary view.
% Our results also show robustness to missing views (see supplementary material).

% \todo[inline]{discuss missing views here?}
% \todo[inline]{refer to extra experiments in supplementary?}

\section{Conclusion}
We proposed an approach that leverages the representational power of conventional deep CNNs and exploits the finite nature of the multiple views to design a group convolutional network that performs an exact equivariance in discrete groups, most importantly the icosahedral group.
We also introduced localized filters and convolutions on homogeneous spaces in this context.
Our method enables joint reasoning over all views as opposed to traditional view-pooling,
and is shown to surpass the state of the art by large margins in several 3D shape retrieval benchmarks.

\section{Acknowledgments}
We are grateful for support through the following grants: NSF-IIP-1439681 (I/UCRC), NSF-IIS-1703319, NSF MRI 1626008, ARL RCTA W911NF-10-2-0016, ONR N00014-17-1-2093, ARL DCIST CRA W911NF-17-2-0181, the DARPA-SRC C-BRIC, and by Honda Research Institute.

% \begin{table}
% \begin{center}
% \begin{tabular}{|l|c|}
% \hline
% Method & Frobnability \\
% \hline\hline
% Theirs & Frumpy \\
% Yours & Frobbly \\
% Ours & Makes one's heart Frob\\
% \hline
% \end{tabular}
% \end{center}
% \caption{Results.   Ours is better.}
% \end{table}

% \newpage
% \listoftodos

{\small
\bibliographystyle{ieee_fullname}
\bibliography{main}
}
\clearpage
\appendix
\section*{Supplementary material: Equivariant Multi-View Networks}
We divide this material in ``Extended experiments'', ``Proofs'', ``Implementation details'', and ``Visualization''.
Numbering and citations follow the main text.

\section{Extended experiments}
We include an ablation experiment and further results for ModelNet, SHREC'17 large scale retrieval, comparison with RotationNet \cite{rotat}, and Matterport3D scene classification.
\subsection{Ablation}
We run an experiment to compare effects of (1) filter support size, (2) number of G-Conv layers, and (3) missing views. % \todo{dset mode?}
We evaluate on rotated ModelNet40 with ``Ours-60'' model as baseline that has with 9 elements in the support, 3 G-Conv layers and all 60 views.

When considering less than 60 views, we introduce view dropout during training where a random number (between 1 and 30) of views is selected for every mini-batch.
This improves robustness to missing views.
During test, a fixed number of views is used.
Table~\ref{tab:ablation} shows the results.
As expected, we can see some decline in performance with fewer layers and smaller support, which reduces the receptive field at the last layer.
Our method is shown to be robust to missing up to 50\% of the views, with noticeable drop in performance when missing 80\% or more.

In the second ablation experiment, we investigate the performance as the assumption that the views can be associated with group elements is broken.
We perturb the camera pose for each view with randomly sampled rotations given some standard deviation.
The model is trained once on ModelNet40 (aligned) and tested under different levels of perturbations.
Table~\ref{tab:jitter} shows the results.

% \todo[inline]{compare 60x1, 20x3, 12x5}
% \todo[inline]{try w/ resnet34, resnet50}

% | 03-28 02:15 | mvgcnn60-m40ti3d-r18-3l-6s-256-3     | 84.55 (84.70)   |    90.63 |   88.62 |    81.90 |   77.05 | 0.3440 |

\begin{table}[h]
  \centering
  \begin{tabular}{cccccc}
    \toprule
    support & layers & views & pretrained & acc   & mAP   \\
    \midrule
    9       & 3         & 60       &   yes         & 91.00 & 82.61 \\
    6       & 3         & 60       &   yes         & 90.63 & 81.90 \\
    3       & 3         & 60       &   yes         & 89.74 & 80.49 \\
    9       & 2         & 60       &   yes         & 91.00 & 81.47 \\
    9       & 1         & 60       &   yes         & 90.88 & 79.59 \\
    9       & 3         & 30       &   yes         & 89.50 & 79.20 \\
    9       & 3         & 10       &   yes         & 88.32 & 74.65 \\
    9       & 3         & 5        &   yes         & 82.77 & 64.88 \\
    9       & 3         & 60       &   no          & 87.40 & 70.44 \\
    \bottomrule
  \end{tabular}
  \caption{Ablation study on rotated ModelNet40.
  Our best performing model is on the top row.}
  \label{tab:ablation}
\end{table}

\begin{table}[h]
\centering
\begin{tabular}{ccc}
\toprule
std [deg] & acc & mAP\\
\midrule
0 & 93.96 & 89.74\\
5 & 94.20 & 89.34\\
15 & 92.70 & 86.97\\
30 & 88.61 & 81.16\\
45 & 80.98 & 71.61\\
\bottomrule
\end{tabular}
\caption{We perturb the camera pose for each view to gradually break the assumption that they form a group.
  The model is trained with no perturbations and tested under different levels of perturbations.}
\label{tab:jitter}
\end{table}

\subsection{SHREC'17}
We show all metrics for the SHREC'17 large scale retrieval challenge in Table~\ref{tab:shrecext}. % that were omitted in the main text due to space constraints.
% We also show Precision (P), Recall (R), F-score (F1), mean average precision (mAP) and normalized discounted cumulative gain (G), where N is the number of retrieved elements.
% \todo[inline]{also include classification results}
% \todo[inline]{show retrieved samples}

\begin{table*}[ht!]
  \centering
    \begin{tabular}{@{}llrrrrrrrrrrrr@{}}
      \toprule
                                               &           & \phantom{} & \multicolumn{5}{c}{micro} & \phantom{A} & \multicolumn{5}{c}{macro}                                                                        \\
      \cmidrule{2-2} \cmidrule{4-8} \cmidrule{10-14}
                  Method                       & score     &            & P@N                       & R@N         & F1@N      & mAP       & G@N       &  & P@N       & R@N       & F1@N      & mAP       & G@N       \\
      \midrule
                  RotatNet~\cite{rotat}        & 67.8      &            & 81.0                      & 80.1       & 79.8      & 77.2      & 86.5      &  & 60.2      & 63.9      & 59.0      & 58.3      & 65.6      \\
                  ReVGG~\cite{savva2017shrec}  & 61.8      &            & 76.5                      & 80.3       & 77.2      & 74.0      & 82.8      &  & 51.8      & 60.1      & 51.9      & 49.6      & 55.9      \\
                  DLAN~\cite{furuya2016deep} & 57.0      &            & 81.8                      & 68.9       & 71.2      & 66.3      & 76.2      &  & 61.8      & 53.3      & 50.5      & 47.7      & 56.3      \\
                  MVCNN-12~\cite{multi}      & 65.1      &            & 77.0                      & 77.0       & 76.4      & 73.5      & 81.5      &  & 57.1      & 62.5      & 57.5      & 56.6      & 64.0      \\
                 MVCNN-M-12                    & 69.1      &            & 83.1                      & 77.9       & 79.4      & 74.9      & 83.8      &  & \it{66.8} & 68.4      & 65.2      & 63.2      & 70.3      \\
                  Ours-12                      & 70.7      &            & 83.1                      & 80.5       & 81.1      & 77.7      & 86.3      &  & 65.3      & 68.7      & 64.8      & 63.6      & 70.8      \\
                  Ours-20                      & 71.4      &            & \it{83.6}                 & \it{80.8}  & \it{81.5} & \it{77.9} & \it{86.8} &  & 66.4      & \it{70.1} & 65.9      & 64.9      & 71.9      \\
                  Ours-60                      & \it{71.7} &            & \bf{84.0}                 & 80.5       & 81.4      & 77.8      & 86.4      &  & \bf{67.1} & \bf{70.7} & \bf{66.6} & \bf{65.6} & \bf{72.3} \\
                  Ours-R-20                    & \bf{72.2} &            & \it{83.6}                 & \bf{81.7}  & \bf{82.0} & \bf{79.1} & \bf{87.5} &  & \it{66.8} & 69.9      & \it{66.1} & \it{65.4} & \bf{72.3} \\
      \midrule
                  DLAN~\cite{furuya2016deep}   & 56.6      &            & \bf{81.4}                 & 68.3       & 70.6      & 65.6      & 75.4      &  & \bf{60.7} & 53.9      & 50.3      & 47.6      & 56.0      \\
      ReVGG~\cite{savva2017shrec}  & 55.7      &            & 70.5                      & \bf{76.9}  & 71.9      & \it{69.6} & 78.3      &  & 42.4      & 56.3      & 43.4      & 41.8      & 47.9      \\
      RotatNet~\cite{rotat}        & 46.6      &            & 65.5                      & 65.2       & 63.6      & 60.6      & 70.2      &  & 37.2      & 39.3      & 33.3      & 32.7      & 40.7      \\
      MVCNN-80~\cite{multi}      & 45.1      &            & 63.2                      & 61.3       & 61.2      & 53.5      & 65.3      &  & 40.5      & 48.4      & 41.6      & 36.7      & 45.9      \\

                 MVCNN-M-60                    & 57.5      &            & 77.7                      & 67.6       & 71.1      & 64.1      & 75.9      &  & 55.7      & 56.9      & 53.5      & 50.9      & 59.7      \\
                  Ours-12                      & 58.1      &            & 76.1                      & 70.0       & 72.0      & 66.4      & 76.7      &  & 54.6      & 55.7      & 52.6      & 49.8      & 58.6      \\
                  Ours-20                      & 59.3      &            & 76.4                      & 70.5       & 72.4      & 66.9      & 77.0      &  & 54.6      & 58.0      & 53.7      & 51.7      & 60.2      \\
                  Ours-60                      & \it{62.1} &            & \it{78.7}                 & 72.9       & \it{74.7} & \it{69.6} & \it{79.6} &  & 57.6      & \it{60.1} & \it{56.3} & \it{54.6} & \it{63.0} \\
                  Ours-R-60                    & \bf{63.5} &            & \it{78.7}                 & \it{75.0}  & \bf{75.9} & \bf{71.8} & \bf{81.1} &  & \it{58.3} & \bf{60.6} & \bf{56.9} & \bf{55.1} & \bf{63.3} \\
      \bottomrule
    \end{tabular}
  \caption{SHREC'17 retrieval results.
    Top block: aligned dataset; bottom: rotated.
    Methods are ranked by the micro and macro mAP average (namely, the ``score'' in the second column).
    We also show Precision (P), Recall (R), F-score (F1), mean average precision (mAP) and normalized discounted cumulative gain (G), where N is the number of retrieved elements.}
  \label{tab:shrecext}
\end{table*}

\subsection{ModelNet}
Since some methods show ModelNet40 results as averages per class instead of the more common average per instance, we include exteded tables with these metrics.
We also present results on rotated ModelNet10.
Tables~\ref{tab:mupext} and \ref{tab:m3dext} shows the results for the aligned and rotated versions, respectively.

% mvgcnn60-m40dup-r18-3l-9s-256-1          | 91.78 (91.78)   |    94.51 |   92.49 |    91.82 |   88.28 |
% mvgcnn60-m10-dup-r18-3l-9s-256-1         | 95.66 (95.67)   |    96.33 |   96.00 |    95.30 |   95.00 |
% mvgcnn60-m40iup-r18-3l-9s-256-1          | 91.59 (91.72)   |    94.69 |   92.56 |    91.42 |   87.71 |
% mvgcnn60-m10-iup-r18-3l-9s-256-1         | 96.53 (96.53)   |    97.46 |   97.34 |    95.74 |   95.58 |
% mvgcnn60-m40tiup-r18-3l-9s-256-1         | 91.28 (91.28)   |    94.36 |   92.40 |    91.04 |   87.30 |
% mvgcnn60-m10-tiup-r18-3l-9s-256-1        | 95.91 (95.91)   |    96.80 |   96.58 |    95.25 |   95.01 |
% mvgcnn60-m40-iup-r18-3l-9s-256-triplet-1 | 92.44 (92.58)   |    94.44 |   92.49 |    93.19 |   89.65 |
% mvgcnn60-m10-iup-r18-3l-9s-256-triplet-1 | 96.76 (96.76)   |    97.02 |   96.97 |    96.59 |   96.46 |

\begin{table*}
  \centering
    \begin{tabular}{lrrrrrrrrr}
      \toprule
                & \multicolumn{4}{c}{M40 (aligned)} & \phantom{abc} & \multicolumn{4}{c}{M10 (aligned)}                              \\
      \cmidrule{2-5} \cmidrule{7-10}
                & acc inst                          & acc cls      & mAP inst & mAP cls &  & acc inst & acc cls & mAP inst & mAP cls \\
      Ours-12   & 94.51                             & 92.49         & 91.82   & 88.28   &  & 96.33    & 96.00    & 95.30   & 95.00   \\
      Ours-20   & 94.69                             & 92.56         & 91.42   & 87.71   &  & 97.46    & 97.34    & 95.74   & 95.58   \\
      Ours-60   & 94.36                             & 92.40         & 91.04   & 87.30   &  & 96.80    & 96.58    & 95.25   & 95.01   \\
      Ours-R-20 & 94.44                             & 92.49         & 93.19   & 89.65   &  & 97.02    & 96.97    & 96.59   & 96.46   \\
      \bottomrule
    \end{tabular}
  \caption{Aligned ModelNet results.
    We include classification accuracy and retrieval mAP per class (cls) and per instance (inst).}
  \label{tab:mupext}
\end{table*}

% | 03-21 09:40 | mvgcnn60-m40d3d-h12-r18-3l-9s-256-h1st-1 | 82.12 (82.37)   |    88.50 |   85.77 |    79.58 |   74.64 | 0.4241 |
% | 03-21 03:45 | mvgcnn60-m40i3d-h20-r18-3l-9s-256-h1st-1 | 83.50 (83.53)   |    89.98 |   87.65 |    80.73 |   75.65 | 0.3988 |
% | 03-21 05:59 | mvgcnn60-m40ti3d-r18-3l-9s-256-1         | 85.22 (85.22)   |    91.00 |   89.24 |    82.61 |   78.02 | 0.3338 |
% | 03-21 07:10 | mvgcnn60-m40ti3d-r18-3l-9s-256-triplet-1 | 88.24 (88.24)   |    91.08 |   88.94 |    88.57 |   84.37 | 0.3461 |

% | 03-27 21:09 | mvgcnn60-m10ti3d-r18-3l-9s-256-1         | 90.53 (90.98)   |    92.83 |   92.80 |    88.47 |   88.02 | 0.1990 |
% | 03-27 21:02 | mvgcnn60-m10ti3d-r18-3l-9s-256-triplet-1 | 92.55 (92.78)   |    93.05 |   93.08 |    92.07 |   91.99 | 0.1960 |
% | 03-27 23:18 | mvgcnn60-m10d3d-h12-r18-3l-9s-256-h1st-1 | 89.11 (89.14)   |    91.89 |   91.54 |    86.93 |   86.08 | 0.2951 |
% | 03-27 23:16 | mvgcnn60-m10i3d-h20-r18-3l-9s-256-h1st-1 | 89.72 (89.72)   |    92.60 |   92.35 |    87.27 |   86.65 | 0.2607 |

\begin{table*}
  \centering
    \begin{tabular}{lrrrrrrrrr}
      \toprule
                & \multicolumn{4}{c}{M40 (rotated)} & \phantom{abc} & \multicolumn{4}{c}{M10 (rotated)}                                   \\
      \cmidrule{2-5} \cmidrule{7-10}
                & acc inst                          & acc cls       & mAP inst cls & mAP cls &  & acc inst & acc cls & mAP inst & mAP cls \\
      Ours-12   & 88.50                             & 85.77         & 79.58        & 74.64   &  & 91.89    & 91.54   & 86.93    & 86.08   \\
      Ours-20   & 89.98                             & 87.65         & 80.73        & 75.65   &  & 92.60    & 92.35   & 87.27    & 86.65   \\
      Ours-60   & 91.00                             & 89.24         & 82.61        & 78.02   &  & 92.83    & 92.80   & 88.47    & 88.02   \\
      Ours-R-20 & 91.08                             & 88.94         & 88.57        & 84.37   &  & 93.05    & 93.08   & 92.07    & 91.99   \\
      \bottomrule
    \end{tabular}
    \caption{Rotated ModelNet results.
    We include ModelNet10 and classification accuracy and retrieval mAP per class (cls) and per instance (inst).}
  \label{tab:m3dext}
\end{table*}

\subsection{Comparison with RotationNet}
We provide further comparison against RotationNet \cite{rotat}.
While RotationNet remains SoTA on aligned ModelNet classification, our method is superior on all retrieval benchmarks.
We also outperform RotationNet on more challenging classification taks: rotated and aligned ShapeNet, and rotated ModelNet.
Table~\ref{tab:rotat} shows the results.

\begin{table}
\small
% \begin{tabular}{@{}llllllllllll@{}}
\begin{tabular}{@{}lllllllll@{}}
\toprule
  %             & \multicolumn{2}{c}{M40 (al)} & \phantom{a}  & \multicolumn{2}{c}{M40 (rot)} & \phantom{a} & \multicolumn{2}{c}{S17 (al)} & \phantom{a} & \multicolumn{2}{c}{S17 (rot)}  \\
  % \cmidrule{2-3}  \cmidrule{5-6} \cmidrule{8-9} \cmidrule{11-12}
  %             & acc            & mAP                        &  & acc            & mAP            &  & acc   & score           &  & acc   & score           \\
  % \midrule
  % RotNet        & \textbf{97.37} & 93.00          &  & 80.0           & 74.20              &  & 85.39    & 67.8          &  & 77.37     & 46.6          \\
  % Ours   & 94.67 & \textbf{93.56}  &  & \textbf{91.08} & \textbf{88.57} &  & \textbf{89.15} & \textbf{72.2} &  & \textbf{85.93} & \textbf{63.5} \\
% we remove M40 (rot) to save space as it's also on the main paper.
& \multicolumn{2}{c}{M40 (al)} & \phantom{.}  & \multicolumn{2}{c}{S17 (al)} & \phantom{.} & \multicolumn{2}{c}{S17 (rot)}  \\
  \cmidrule{2-3}  \cmidrule{5-6} \cmidrule{8-9}
              & acc            & mAP                                   &  & acc   & mAP           &  & acc   & mAP           \\
  \midrule
  RotNet        & \textbf{97.37} & 93.00          &    & 85.39    & 67.8          &  & 77.37     & 46.6          \\
  Ours   & 94.67 & \textbf{93.56}  &  &  \textbf{89.15} & \textbf{72.2} &  & \textbf{85.93} & \textbf{63.5} \\
\bottomrule
\end{tabular}

\caption{Classification accuracy (acc) and retrieval (mAP) comparison against RotationNet.
  Results for aligned (al) and rotated (rot) datasets, and for the SHREC'17 split of ShapeNet (S17).
  The mAP for SHREC'17 is the average between micro and macro (score).
}
\label{tab:rotat}
\end{table}

\subsection{Scene classification}
\begin{figure*}[h]
  \centering
  \includegraphics[width=0.8\linewidth]{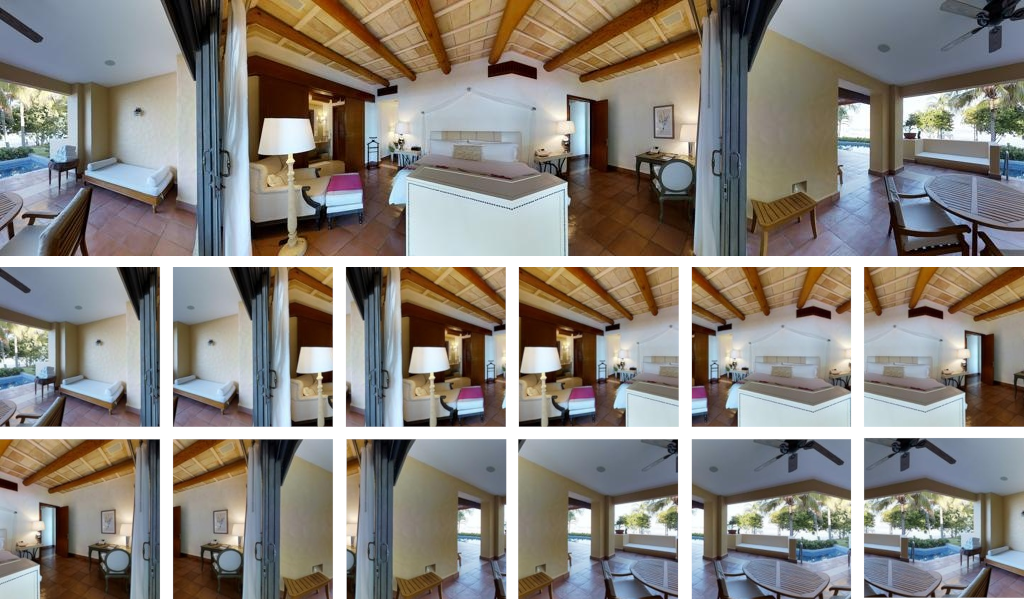}
  \caption{Top: original input from MatterPort3D~\cite{chang17_matter} scene classification task.
    Bottom: our set of 12 overlapping views.}
\label{fig:matterport}
\end{figure*}

We show examples of original input and our 12 overlapping views in Figure~\ref{fig:matterport}.
The complete set of results in the same format as~\cite{chang17_matter} is shown in Table~\ref{tab:matterportcls}.

%\todo[inline]{include dataset samples}

% mvgcnn12-mat12rgb-r18-3l-6s-256-3
% acc per class=[27.9, 16.4, 33.3, 11.4, 51.1, 41.3, 80.4, 75.8, 79.0, 72.5, 82.9, 73.5]
% 27.9& 16.4& 33.3& 11.4& 51.1& 41.3& 80.4& 75.8& 79.0& 72.5& 82.9& 73.5 \\

% mvcnn12-mat12rgb-r18-3l-3
% acc per class=[18.0, 16.4, 23.8, 8.6, 46.7, 37.1, 84.1, 73.3, 81.0, 78.2, 81.7, 73.8]
% 18.0& 16.4& 23.8& 8.6& 46.7& 37.1& 84.1& 73.3& 81.0& 78.2& 81.7& 73.8\\

\begin{table*}
  \begin{center}
    \scriptsize
    \begin{tabular}{lrrrrrrrrrrrrr}
      \toprule
                                   & avg.      & office    & lounge    & family room & entryway  & dining room & living room & stairs    & kitchen   & porch     & bathroom  & bedroom   & hallway   \\
      \midrule
      single~\cite{chang17_matter} & 33.3      & 20.3      & \bf{21.7} & 16.7        & 1.8       & 20.4        & 27.6        & 49.5      & 52.1      & 57.4      & 44.0      & 43.7      & 44.7      \\
      pano~\cite{chang17_matter}   & 41.0      & \it{26.5} & 15.4      & 11.4        & 3.1       & 27.7        & 34.0        & 60.6      & 55.6      & 62.7      & 65.4      & 62.9      & 66.6      \\
      MVCNN-M-12                   & \it{51.9} & 18.0      & \it{16.4} & \it{23.8}   & \it{8.6}  & \it{46.7}   & \it{37.1}   & \bf{84.1} & \it{73.3} & \bf{81.0} & \bf{78.2} & \it{81.7} & \bf{73.8} \\
      Ours-12                      & \bf{53.8} & \bf{27.9} & \it{16.4} & \bf{33.3}   & \bf{11.4} & \bf{51.1}   & \bf{41.3}   & \it{80.4} & \bf{75.8} & \it{79.0} & \it{72.5} & \bf{82.9} & \it{73.5} \\
      \bottomrule
    \end{tabular}
  \end{center}
  \caption{Matterport3D panoramic scene classification extended results.}
  \label{tab:matterportcls}
\end{table*}

\section{Proofs}
We demonstrate the equivariance of operations, and show that MVCNN~\cite{multi} is a special case of our method.
\subsection{Equivariance of G-Conv, H-Conv, H-Corr}
We demonstrate here the equivariance of the main operations used in our method: group convolution, homogeneous space convolution and correlation.
We assume a compact group $G$ and one of its homogeneous spaces $\mathcal{X}$.
$\mathcal{T}_{k}$  is the action of $k\in G$.
Let us start with G-Conv (\ref{eq:gconv}), where \fun{f,h}{G}{\R}:
\begin{align*}
  (\mathcal{T}_{k}f*h)(y)
%&=\int_{g\in G}\mathcal{T}_{k}f(g)h(g^{-1}y)\,dg
&=\int_{g\in G}f(k^{-1}g)h(g^{-1}y)\,dg\\
&=\int_{g\in G}f(l)h((kl)^{-1}y)\,dl\\
&=\int_{g\in G}f(l)h((l^{-1}k^{-1}y)\,dl\\
&=(f*h)(k^{-1}y)\\
&=\mathcal{T}_{k}(f*h)(y).
\end{align*}           %G-conv
For H-Conv (\ref{eq:hconv}), where \fun{f,h}{\mathcal{X}}{\R}, we have:
\begin{align*}
  (\mathcal{T}_{k}f* h)(y)
% &= \int_{g\in G}\mathcal{T}_{k}f(g\eta)h(g^{-1}y)\,dg\\
&= \int_{g\in G}f(k^{-1}g\eta)h(g^{-1}y)\,dg\\
&= \int_{g\in G}f(l\eta)h((kl)^{-1}y)\,dl\\
&= \int_{g\in G}f(l\eta)h(l^{-1}k^{-1}y)\,dl\\
&=(f*h)(k^{-1}y)\\
&=\mathcal{T}_{k}(f*h)(y).
\end{align*}              %H-conv
Finally, for H-Corr (\ref{eq:hcorr}), where \fun{f,h}{\mathcal{X}}{\R}, we have:
\begin{align*}
(\mathcal{T}_{k}f \star h)(g)
% &=\int_{x\in \mathcal{X}}\mathcal{T}_{k}f(gx)h(x)\,dx\\
&=\int_{x\in \mathcal{X}}f(k^{-1}gx)h(x)\,dx\\
&=(f \star h)(k^{-1}g)\\
&=\mathcal{T}^{'}_{k}(f \star h)(y).
\end{align*}
Note that, in this case, $\mathcal{T}^{'}_{k}$ is not equal $\mathcal{T}_{k}$ because inputs and outputs are in different spaces.

\subsection{MVCNN is a special case}
%\todo[inline]{update notation: filters are $h_{ij}(g)$}
%Now we show that we can use EMVN to achieve the same result as MVCNN~\cite{multi} by designing the group filter $W(O,I,g)$ ($O$ denotes the output channel, $I$ denotes the input channel and $g$ denotes the element in group) as:
Now we show that our model can replicate MVCNN~\cite{multi} by fixing the filters $h_{ij}(g)$, where $i,j$ denote the output and input channel and $g$ denotes the element in group:
%\begin{equation}
%W(O,I,g)=\left\{
%\begin{array}{rcl}
%1& & {I=O}\quad \text{and} \quad {g=e}\\
%0 & & \text{else}
%\end{array} \right.
%\end{equation}

\begin{equation}
h_{ij}(g)=\left\{
\begin{array}{rcl}
1& & {i=j}\quad \text{and} \quad {g=e}\\
0 & & \text{else}
\end{array} \right.
\end{equation}

Combining with group correlation (the result can also be achieved by group convolution), we show that\\
%\begin{align*}
%(F\star W)_{(O,h)}&=\sum_I\sum_{g \in G}F(I,hg)W(O,I,g)\\
%&=\left\{
%\begin{aligned}
%&F(O,h) && 1\leq O\leq N_{I}\\
%&\quad 0  && O>N_{I}
%\end{aligned} \right.,
%\end{align*}
\begin{align*}
(f\star h)_{i}(k) &= \sum_{j=1}^{c_1} \sum_{g \in G}f_j(kg)h_{ij}(g)\\
&=\left\{
\begin{aligned}
&f_i(k) && 1\leq i\leq c_{i}\\
&\quad 0  && i>c_{i}
\end{aligned} \right.,
\end{align*}
where $c_{i}$ is the number of input channels. In this way, the input is ``copied'' into the output and the our model produces the exact same descriptor as an MVCNN with late pooling after the last layer.

\begin{figure*}[t!]
  \centering
  \includegraphics[width=\linewidth]{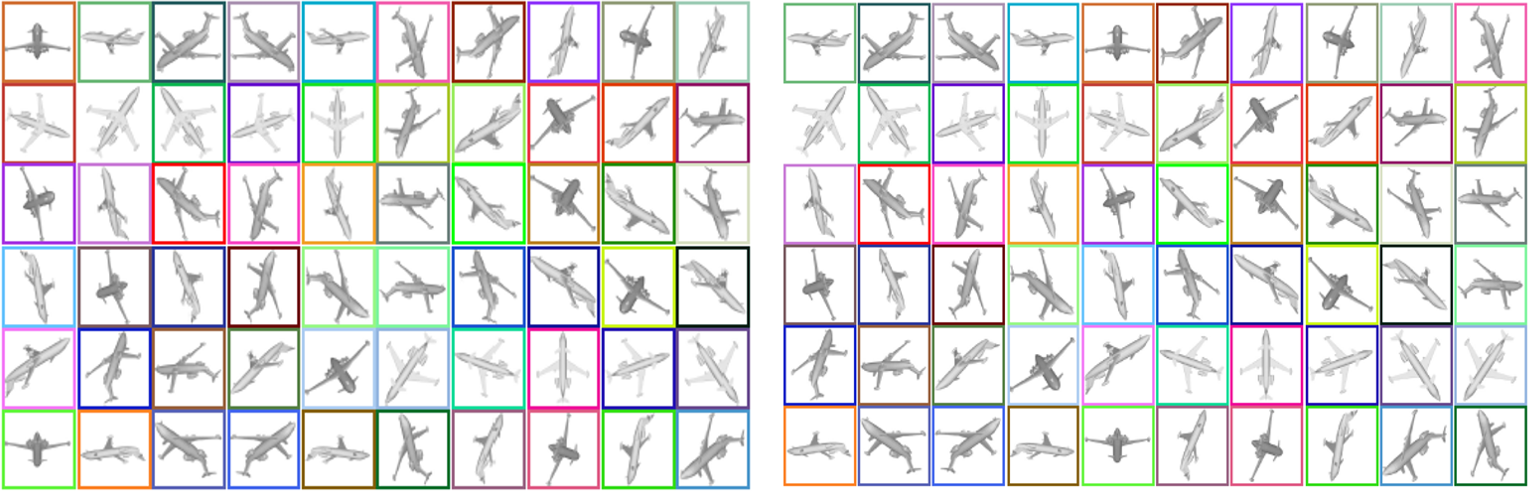}
  \caption{Equivariance of view configurations to $\I$.
    The views on the left and right are obtained from 3D shapes separated by a 72 deg rotation in the discrete group.
    We mark corresponding views before and after rotation with same border color.
    Notice the five first views in the second row -- the axis of rotation is aligned with their optical axis; the rotation effect is a shift right of one position for these views.
    It is clear that when $g \in \I$ is applied to the object, the views are correspondingly permuted in the order given by the Cayley table, showing that the mapping from 3D shape to view set is equivariant.}
    \label{fig:equi60}
\end{figure*}

\section{Implementation details}
We include details about our triplet loss implementation and about our procedure for visualization of discrete rotation groups and their homogeneous spaces.
\subsection{Triplet loss}
We implement a simple triplet loss.
During training, we keep a set containing the descriptors for the last seen instance of each class, $Z=\{z_i\}$, where $i$ is the class label.
For each entry in the mini-batch, let $c$ be the class and $z$ its descriptor.
We take the descriptor in $Z$ of the same class as a positive example ($z_c$), and chose the hardest between all the others in the set as the negative: $z_n = \text{argmin}_{z_i\in Z,\,i\neq c}(d(z_i,z))$, where $d$ is a distance function.
The contribution of this entry to the loss is then,
\begin{equation}
  \mathcal{L} = \max(d(z,z_c) - d(z,z_n) + \alpha, 0),
\end{equation}
where $\alpha$ is a margin.
We use $\alpha=0.2$ and $d$ is the cosine distance.
Note that this method is only used in the ``Ours-R'' variations of our method.

\subsection{Feature map visualization}
Our features are functions on a subgroup of the rotation group \SO(3).
Since \SO(3) is a 3-manifold (which can be embedded in $\R^5$), visualization is challenging.
As we operate on the discrete subgroup of 60 rotations, we choose a solid with icosahedral symmetry and 60 faces as a proxy for visualization -- the pentakis dodecahedron, which is the dual of the truncated icosahedron (the ``soccer ball'' with 60 vertices).

We associate the color of each face with the feature vector at that element of the group.
Since the vector is high-dimensional (usually 256 or 512-D), we use PCA over all feature vectors in a layer (or groups of channels in a layer) and project it into the 3 principal components that can be associated with an RGB value.
The same idea is applied to visualize functions on the homogeneous spaces, where the dodecahedron and icosahedron are used as proxies.

\section{Visualization}
We visualize the icosahedral group properties and include more examples of our equivariant feature maps.
\subsection{Icosahedral group}
\begin{figure*}[htbp!]
\centering
  \includegraphics[width=\textwidth]{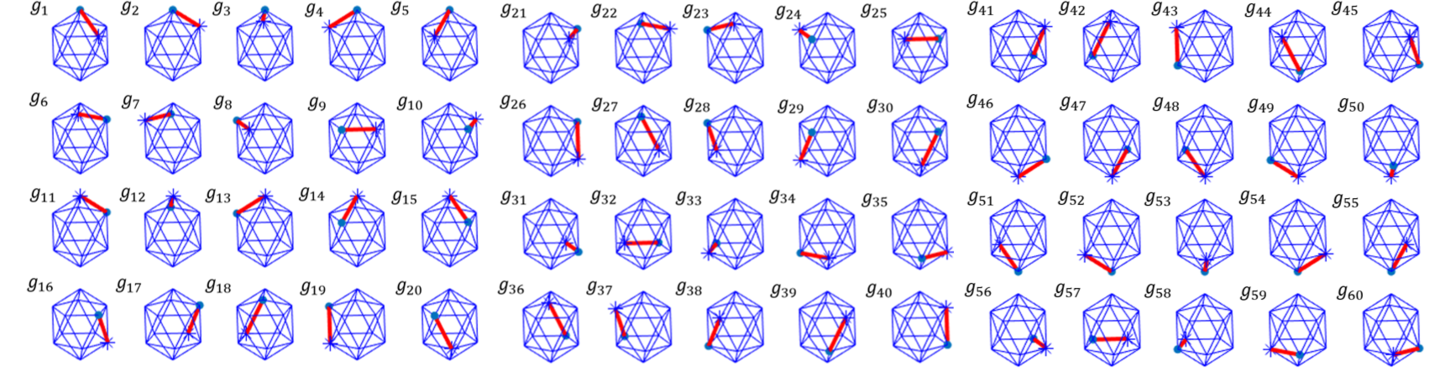}
  \caption{The 60 rotations of the icosahedral group $\I$.
    We consider $g_1$ the identity, highlight one edge, and show how each $g_i \in \I$ transforms the highlighted edge.}
  \label{rotation60}
\end{figure*}

\begin{figure}[H]
  \centering
  \includegraphics[width=\linewidth]{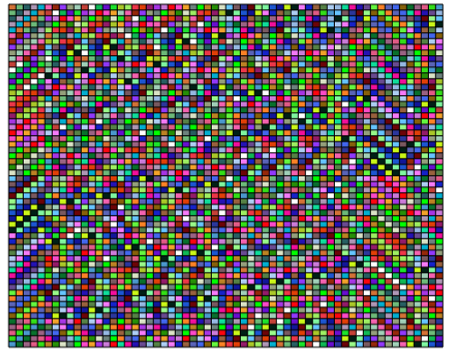}
  % }
  \caption{Cayley table for the icosahedral group $\I$.
    We can see that $\I$ is non-abelian, since the table is not symmetric.}
  \label{Caylay2}
\end{figure}

The icosahedral group $\I$ is the group of symmetries of the icosahedron, which consists of 60 rotations, as visualized in Figure~\ref{rotation60}.
We show how the equivariance manifests as permutation when rendering multiple views of an object according to the group structure in Figure~\ref{fig:equi60}.
Figure~\ref{Caylay2} shows the Cayley Table for $\I$; note that the color assigned for each group element matches the color in Figure~\ref{fig:equi60}.

% \begin{figure}[H]
%   \centering
%  %\subfigure[Caylay Table for group $\boldsymbol{C_{12}}$]{\label{Caylay1}
%  \includegraphics[width=\linewidth]{figures/caylay121.png}

%  \caption{Cayley tables for $\boldmath{C}_{12}$. We can see that $\boldmath{C}_{12}$ is an abelian group for that $\forall g_i, g_j \in G$, $g_ig_j=g_jg_i$}
%  \label{Caylay1}
%     %}
%   %\subfigure[Caylay Table for group $\boldsymbol{I}$]{\label{Caylay2}
% \end{figure}

\subsection{Feature maps}
\begin{figure}[h]
  \centering
  \includegraphics[width=\linewidth]{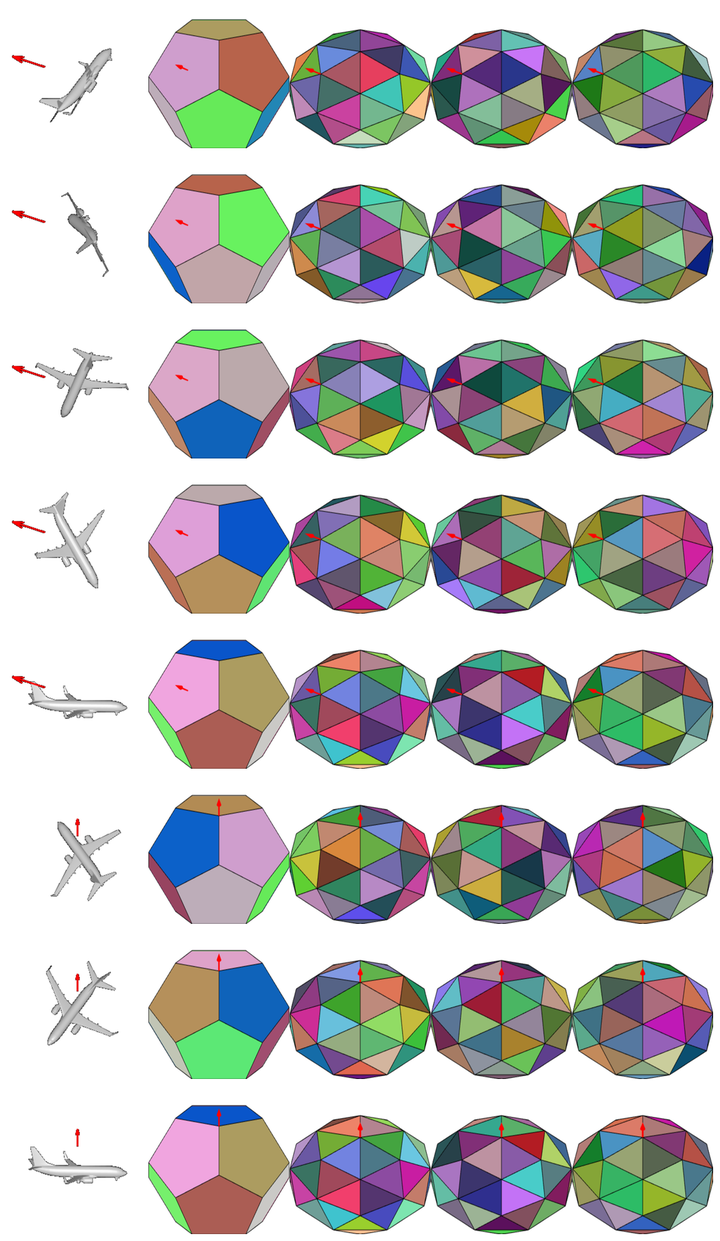}
  \caption{Feature maps with 12 input views.
  See \texttt{animation12.gif} for animated version.}
  \label{fig:fmapsdodec}
\end{figure}

\begin{figure}[h]
  \centering
  \includegraphics[width=\linewidth]{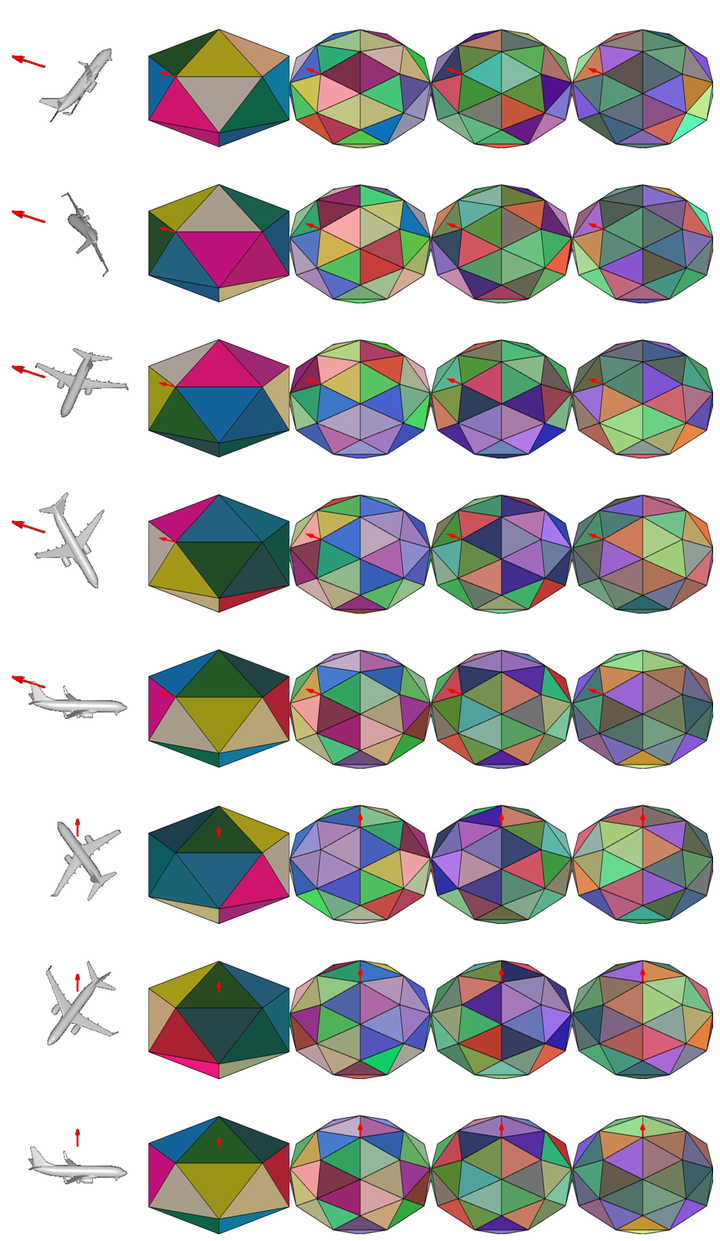}
  \caption{Feature maps with 20 input views.
    See \texttt{animation20.gif} for animated version.}
  \label{fig:fmapsico}
\end{figure}

\begin{figure}[h]
  \centering
  \includegraphics[width=\linewidth]{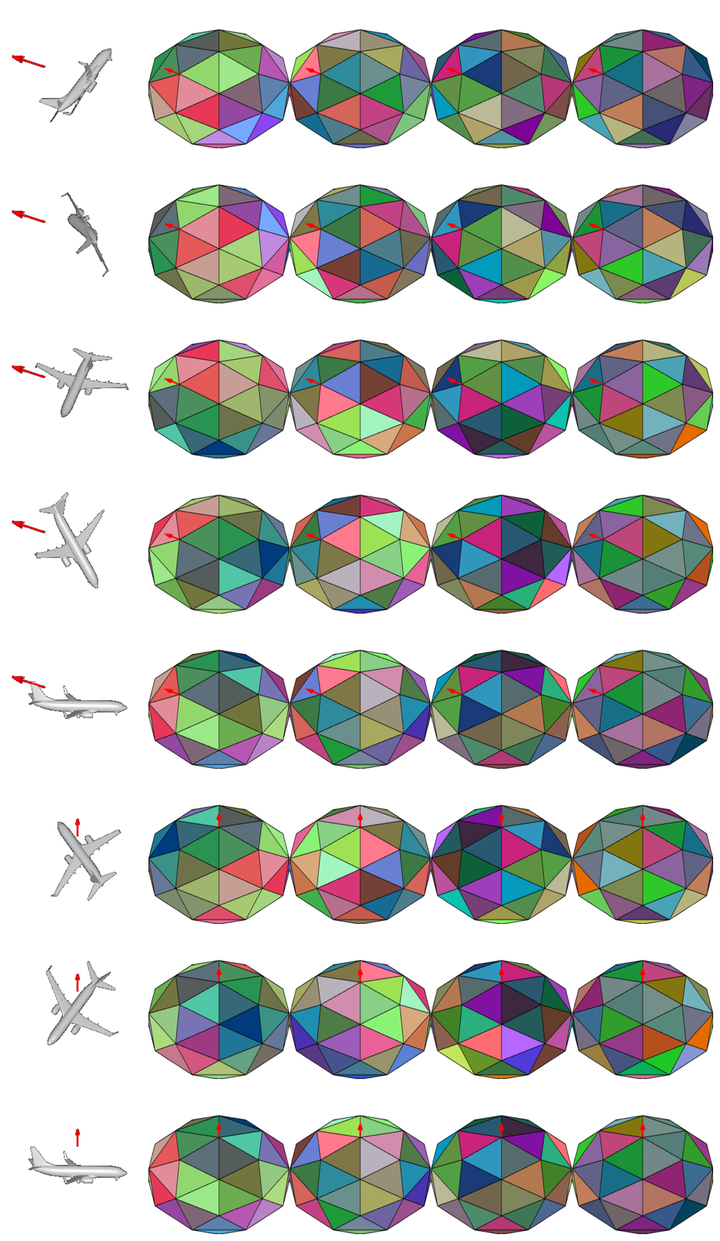}
  \caption{Feature maps with 60 input views.
    See \texttt{animation60.gif} for animated version.}
  \label{fig:fmapspentakis}
\end{figure}

We visualize more examples of our equivariant feature maps in Figures~\ref{fig:fmapsdodec}, \ref{fig:fmapsico}, \ref{fig:fmapspentakis}.
Each figure shows 8 different input rotations, the first 5 are from a subgroup of rotations around one axis with 72 deg spacing, the other 3 are from other subgroup with 120 deg spacing.
We show the axis of rotation in red.
The first column is a view of the input, the second is the initial representation on the group or H-space, and the other 3 are features on each G-CNN layer.

Our method is equivariant to the 60-element discrete rotation group even with only 12 or 20 input views.
In Figure~\ref{fig:fmapsdodec} we take only 12 input views, giving initial features on the H-space represented by faces of the dodecahedron.
Note that the 5 first rotations in this case are in-plane for the views corresponding to the axis of rotation.
Due to our procedure described in Section \ref{sec:fewerviews}, this gives an invariant descriptor which can be visualized as the face with constant color.
Similarly, in Figure~\ref{fig:fmapsico}, we take 20 views and the invariant descriptor can be seen in the last 3 rotations.

Equivariance is easily visualized on faces neighboring the axis of rotation.
For the dodecahedron, we can see cycles of 5 when the axis is on one face and cycles of 3 when the axis is on one vertex.
For the icosahedron, we can see cycles of 3 when the axis is on one face and cycles of 5 when the axis is on one vertex.
For the pentakis dodecahedron (Figure \ref{fig:fmapspentakis}), we can see groups of 5 cells that shift one position when rotation is of 72 deg and groups of 6 cells that shift two positions when rotation is of 120 deg.

\end{document}